\documentclass[letterpaper, 10 pt, conference]{ieeeconf}  

\IEEEoverridecommandlockouts                              

\usepackage{xcolor}
\usepackage{array}
\usepackage{url}
\usepackage{graphicx}
\usepackage{amsmath}
\usepackage{subcaption}
\usepackage{adjustbox}
\usepackage{makecell}
\usepackage{booktabs}
\usepackage{tabularx}
\usepackage{blindtext}
\usepackage[most]{tcolorbox}

\let\labelindent\relax
\usepackage{enumitem} 
\setlist{nolistsep} 
\setlist{leftmargin=3mm}

\title{\LARGE \bf
Supporting Productivity Skill Development in College Students through Social Robot Coaching: A Proof-of-Concept
}

\author{Himanshi Lalwani$^{1}$ and Hanan Salam$^{1}$
\thanks{$^{1}$SMART Lab, Department of Computer Science, New York University, Abu Dhabi
        {\tt\small {hanan.salam@nyu.edu, himanshi.lalwani@nyu.edu}}}
}

\begin{document}

\maketitle
\thispagestyle{empty}
\pagestyle{empty}

\begin{abstract}

College students often face academic challenges that hamper their productivity and well-being. Although self-help books and productivity apps are popular, they often fall short. Books provide generalized, non-interactive guidance, and apps are not inherently educational and can hinder the development of key organizational skills. Traditional productivity coaching offers personalized support, but is resource-intensive and difficult to scale. In this study, we present a proof-of-concept for a socially assistive robot (SAR) as an educational coach and a potential solution to the limitations of existing productivity tools and coaching approaches. The SAR delivers six different lessons on time management and task prioritization. Users interact via a chat interface, while the SAR responds through speech (with a toggle option). An integrated dashboard monitors progress, mood, engagement, confidence per lesson, and time spent per lesson. It also offers personalized productivity insights to foster reflection and self-awareness. We evaluated the system with 15 college students, achieving a System Usability Score of 79.2 and high ratings for overall experience and engagement. Our findings suggest that SAR-based productivity coaching can offer an effective and scalable solution to improve productivity among college students.
\end{abstract}

\section{INTRODUCTION}
Transitioning to college is a critical period marked by both exciting opportunities and significant challenges. Many students leave the comfort of home for the first time and suddenly find themselves responsible for regulating their sleep, daily routines, and overall lifestyle, all while adapting to a new and often overwhelming environment \cite{duffy2020predictors}. Alongside these personal adjustments, college students face a host of academic stressors, ranging from heavy course loads and intense studying to classroom competition, financial concerns, and familial pressures, which can severely impact their overall productivity and academic performance \cite{barbayannis2022academic, pascoe2020impact}. These academic challenges are closely linked to mental health concerns; research shows that approximately 20\% of college students worldwide develop mental health disorders during their first year, with first-generation students and minority groups reporting even greater obstacles \cite{auerbach2016mental, hyseni2024overcoming}. For example, first-generation students have noted statistically significant higher incidences of issues such as balancing job and family responsibilities, lacking essential study skills, and experiencing heightened feelings of stress and depression \cite{stebleton2012breaking}, while women and non-binary students also report increased stress levels compared to their peers \cite{verma2011gender, budge2020minority}. Hence, enhancing productivity through tailored strategies becomes crucial not only for academic success, but also for improving overall well-being of college students.

Building on these challenges, many productivity applications and self-help books have emerged to help students manage their time, reduce stress, and stay focused. Apps like Google Calendar\footnote{https://calendar.google.com/}, Notion\footnote{https://notion.so/}, and Forest\footnote{https://www.forestapp.cc/} offer convenient digital support for organizing tasks and maintaining focus. However, while these tools assist with structure, they are not inherently educational—they do not teach the cognitive or behavioral strategies needed to build sustainable productivity habits. Relying too heavily on these apps can not only limit opportunities to develop core organizational skills but also result in wasted time due to constant app switching \cite{habit10x_limitations, ringcentral_connected_workplace}. In contrast, productivity books offer educational value by providing deeper insights into habit formation and productivity-related challenges. Yet, these books often follow a one-size-fits-all approach, lacking personalization or contextual relevance \cite{bergsma2008self}. Some may offer outdated or unscientific advice, and without interactive support, users are left to independently interpret, select, and implement strategies \cite{selfhelp_myths_2001, rosen1987self}. This places a high burden of responsibility and self-discipline on the reader, which can be challenging for students already struggling with executive function.

Given the shortcomings of traditional productivity applications and self-help books, coaching has become a promising avenue for providing personalized support to students. Student success centers have begun to appear across university campuses, aiming to deliver a range of services including tutoring, mentoring, academic coaching, and supplemental instruction \cite{tsymbaliuk2024academic}. When delivered effectively, these programs can boost essential skills such as self-efficacy, self-awareness, self-regulation, motivation, and goal-setting, all of which are crucial for sustaining productivity and academic success \cite{saethern2022students, mouganie2022college}. However, intensive support programs often require significant resources, making them difficult to scale on a broader level \cite{mouganie2022college}. In contrast, minimal support approaches, such as virtual coaching via automated emails and text messages, are more scalable but may not offer the comprehensive, individualized assistance needed to fully address students' academic challenges \cite{canaan2023keep}.

Socially Assistive Robots (SARs) \cite{feil2005defining} offer a promising alternative to address the limitations of existing productivity tools and resources. Several SAR applications have been developed for college students in recent years. For example, \cite{lalwani2025study} created a productivity assistant that helps college students with ADHD design daily schedules and initiate focus work sessions. Similarly, \cite{9223588} developed a robot that delivers positive psychology exercises, while \cite{rice2023effectiveness} introduced a robotic system aimed at reducing stress among students. In another instance, \cite{6891009} designed a SAR to provide verbal encouragement and boost engagement in mathematics education, and \cite{O'connell} developed an in-dorm companion to support college students with ADHD during schoolwork. Beyond academic support, SARs have also been applied in coaching contexts. For example, Haru \cite{10.1145/3610978.3640583} was used for behavior change coaching, and another robotic coach was designed to support motor, social, and cognitive skills training in children with autism \cite{9461769}. Although SARs have been developed for college students and have been applied in coaching contexts, their role in delivering educational coaching specifically aimed at productivity-related soft skills development remains underexplored.

To support natural and engaging interactions for effective coaching, SARs can leverage large language models (LLMs) with prompt engineering to generate context-aware dialogue. Prompt engineering refers to the process of designing input queries or prompts to elicit specific responses from an LLM \cite{marvin2023prompt}. This technique has enabled LLMs to function as behavior change coaches \cite{guyre2024prompt}, management coaches \cite{meywirth2024designing}, and career coaches \cite{renji2025steve}. In academic contexts, LLMs have been used to generate explainable, personalized study plans \cite{chun2025planglow} and to scaffold learning behaviors, such as overcoming procrastination, through personalized prompts \cite{bhattacharjee2024understanding}. However, to the best of our knowledge, LLMs have not been formally evaluated in the context of productivity coaching, nor have they been integrated into SARs for this purpose.

In this study, we introduce a novel approach to productivity coaching using a socially assistive robot tailored specifically for college students. Acting as an educational coach, our LLM-equipped SAR supports the development of key productivity-related soft skills, including time management, task prioritization, and sustained focus. At the start, the coach engages in a conversation to understand the student's productivity-related challenges. Based on this interaction, the system tailors the subsequent interactions to be more relevant. It also recommends a sequence of six targeted lessons on time management and task prioritization. As students progress through these lessons, an interactive dashboard tracks metrics such as overall progress, mood, top engagement lessons, confidence per lesson, and time spent per lesson. Additionally, the system analyses the interactions to display personal productivity insights along with actionable suggestions on the dashboard. By displaying these metrics, the dashboard fosters self-awareness, which is essential for effective skill development. This increased awareness can help students recognize patterns, make informed decisions, and strengthen self-regulation, while also reducing stress and improving resilience \cite{dierdorff2015research, london2023developing, donald2019does}. Our study evaluates whether this proof-of-concept for the SAR-based coaching system can effectively serve as a productivity coach, offering a scalable yet personalized support option for students. 


\section{METHODOLOGY}
\subsection{Robotic Platform} We use QTrobot\footnote{https://luxai.com/product/qtrobot-rd-v2-aiedge/} by LuxAI as a robotic platform for delivering productivity coaching. QTrobot is equipped with a robust text-to-speech engine, enabling clear and natural verbal communication. Additionally, its ability to perform a variety of gestures and express a range of emotions enhances its social presence, making interactions more engaging. These features collectively facilitate a more natural, human-like conversation, positioning QTrobot as an ideal platform for productivity related coaching in a college setting.
\begin{figure}
  \centering
  \includegraphics[scale=0.20]{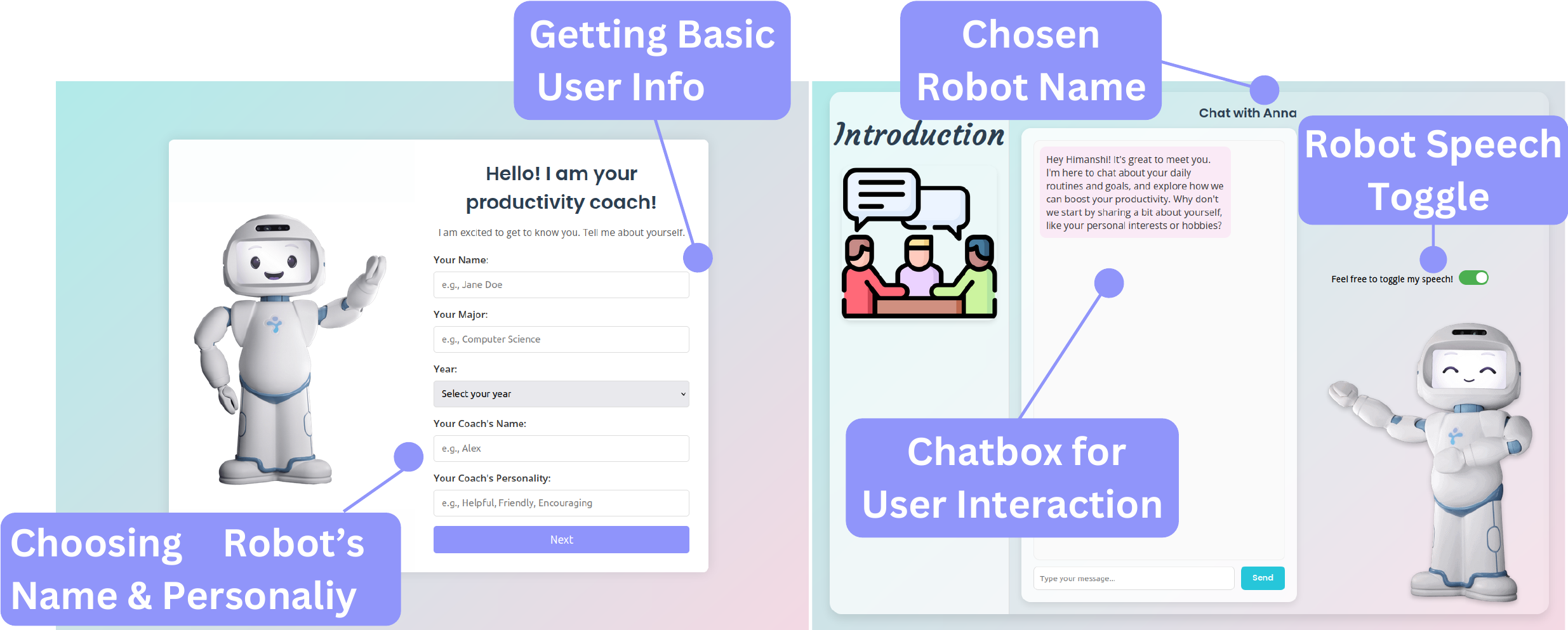}
  \caption{On Left: Home Page. On Right: Introductory Interaction Page.}
  \label{fig:intro}
\end{figure}

\subsection{Robotic System for Productivity Coaching} The robotic coaching system is implemented as a Flask-based web application with an interactive chat interface. Students communicate with the system using text-based chat, with the robot delivering responses verbally by default while also displaying them in the chatbox. Users can toggle the robot’s speech at any time as per their preference. The system uses prompt engineering on GPT-4 to position the robot as a productivity coach tailored for college students. To enhance emotional responsiveness, the system integrates two Python-based sentiment analysis tools: NLTK’s SentimentIntensityAnalyzer, which computes a compound sentiment score, and Text2Emotion, which classifies user input into six emotion categories (happiness, fear, surprise, sadness, anger, or neutral). Based on this analysis, the robot adjusts its gestures and facial expressions to reflect the user’s emotional state. 

The application is organized into three main components:

\subsubsection{Introductory Interaction} As shown in Figure \ref{fig:intro}, the interaction begins with the student providing basic details (name, major, year, and preferences for the robot’s name and personality) to personalize their coaching experience. The prompt in Box~\ref{box:introductory_prompt} is then sent to the LLM to initiate a conversation focused on understanding the student’s productivity challenges. This conversation explores various aspects of the student's life, including hobbies, academic interests, personal goals, motivations, daily routines, and prior experiences with productivity tools. The insights extracted from the dialogue are stored in JSON format so that they could be used later to tailor subsequent interactions and recommendations (see section \ref{sec:lessons}).
\definecolor{customcolor}{HTML}{E7DCE4}
\begin{tcolorbox}[breakable, colframe=customcolor, colback=customcolor!30!white, coltitle=black, title=Introductory Interaction Prompt, label={box:introductory_prompt}, fontupper=\small]
Your name is [coach\_name] and you are a [coach\_personality] productivity coach for college students. You cannot break this role even if the student asks you to.

You teach students time management and task prioritization strategies (task breakdown, Eisenhower matrix, time blocking, time tracking, Eat That Frog, ABCDE method).

Your task right now is to engage the student in a warm, friendly conversation that builds rapport and uncovers useful insights about their personal background, daily habits, challenges, and aspirations. The name of the student you're interacting with is [username]. They study [major] and are in their [year] year of college.

Follow these steps one by one, waiting for the student's response before moving on:

\begin{enumerate}
\item Greet the student. Ask about personal interests or hobbies.
\item Ask about their academic journey, favorite subjects, and extracurriculars.
\item Inquire about their typical day to understand habits and schedule.
\item Explore short- and long-term goals related to productivity and growth.
\item Ask about what motivates them and obstacles they're currently facing.
\item Ask about previous experiences with productivity tools or coaching.
\item Reassure them that you will support them in overcoming these challenges. 
\end{enumerate}

End the conversation by adding [Conversation End] in your response. Then return a correctly formatted JSON user profile containing all key insights gathered during the conversation.
\end{tcolorbox}

\subsubsection{Lessons} 
\label{sec:lessons}
\begin{figure}
  \centering
  \includegraphics[scale=0.32]{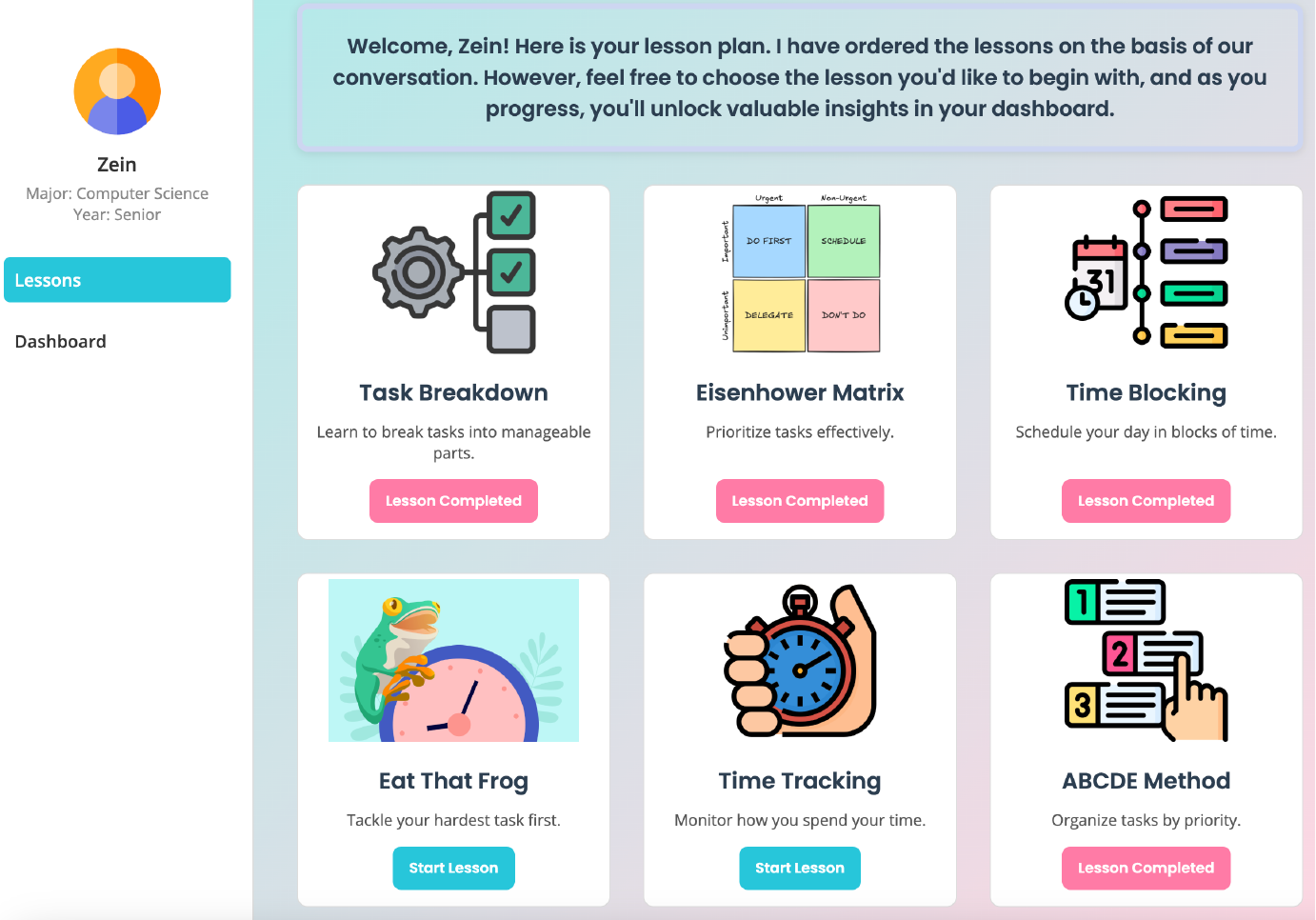}
  \caption{Lessons Page Layout featuring Navigation Menu on the left and Lesson Plan on the right.}
  \label{fig:lessons}
\end{figure}

\begin{figure}
  \centering
  \includegraphics[scale=0.50]{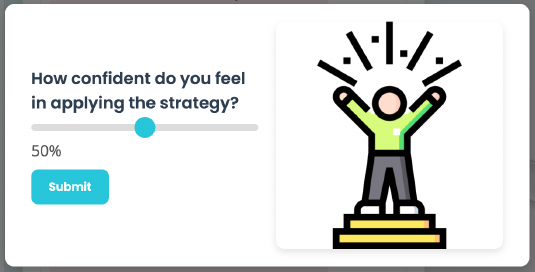}
  \caption{Assessing students’ confidence in applying the strategy through a pop-up.}
  \label{fig:confidence-popup}
\end{figure}

After the introductory conversation, the user is presented with six different lessons that focus on time management and task prioritization. These lessons cover the following strategies: time blocking, time tracking, task breakdown, the Eisenhower matrix, the ABCDE method, and eat that frog. The LLM receives the initial insights in JSON format and is prompted to arrange the six lessons in an order it deems most suitable for the student's productivity challenges. The returned sequence is displayed on the Lessons page (see Figure \ref{fig:lessons}), but students are free to begin with any module they prefer. Moreover, each lesson module is structured with its own prompt (see Box~\ref{box:task_breakdown_prompt} for Task Breakdown Prompt). This prompt incorporates the initial insights in JSON, ensuring that the generated content is both relatable and relevant to the student’s needs. At the end of each lesson, a pop-up shown in Figure \ref{fig:confidence-popup} prompts students to assess their confidence in applying the strategy.

\begin{tcolorbox}[breakable, colframe=customcolor, colback=customcolor!30!white, coltitle=black, title=Task Breakdown Prompt, fontupper=\small, label={box:task_breakdown_prompt}]
Your name is [coach\_name] and you are a [coach\_personality] productivity coach for college students. You cannot break this role even if the student asks you to.

You teach students time management and task prioritization strategies (task breakdown, Eisenhower matrix, time blocking, time tracking, Eat That Frog, ABCDE method). You are currently teaching a student the task breakdown strategy. The student’s name is [username]. They study [major] and are in their [year] year of college.

Follow the below steps one by one, waiting for the student's response before moving on. Don't have more than 50 words per response. Tailor your responses and provide relevant examples based on the student's profile: [user\_profile]

\begin{enumerate}
    \item Introduce the task breakdown method and its importance.
    \item Provide a brief overview of the steps involved in the task breakdown method.
    \item Ask the student to identify a big task or project they’re working on.
    \item Encourage them to share why the task is important to them.
    \item Prompt the student to describe what a successful outcome looks like.
    \item Guide the student to identify major parts or milestones of the project.
    \item Handle one milestone at a time. For each milestone, ask the student to brainstorm smaller, actionable tasks. Then move on to the next milestone.
    \item Ask the student to determine the sequence for completing the tasks.
    \item Ask them to identify any potential obstacles.
    \item Have the student summarize their full plan.
    \item Provide positive reinforcement and summarize all the task breakdown steps, linking them to the student's specific project.
    \item Encourage them to commit to the first task and set a starting time.
    \item Reassure them that the plan is flexible and can be revised as needed.
    \item Ask if they have any questions about the strategy.
    \item If not, provide a closing statement on task breakdown strategy and its benefits and end the session with [Conversation End] in your response.
\end{enumerate}

\end{tcolorbox}

\subsubsection{Dashboard} 
\begin{figure}
  \centering
  \includegraphics[scale=0.31]{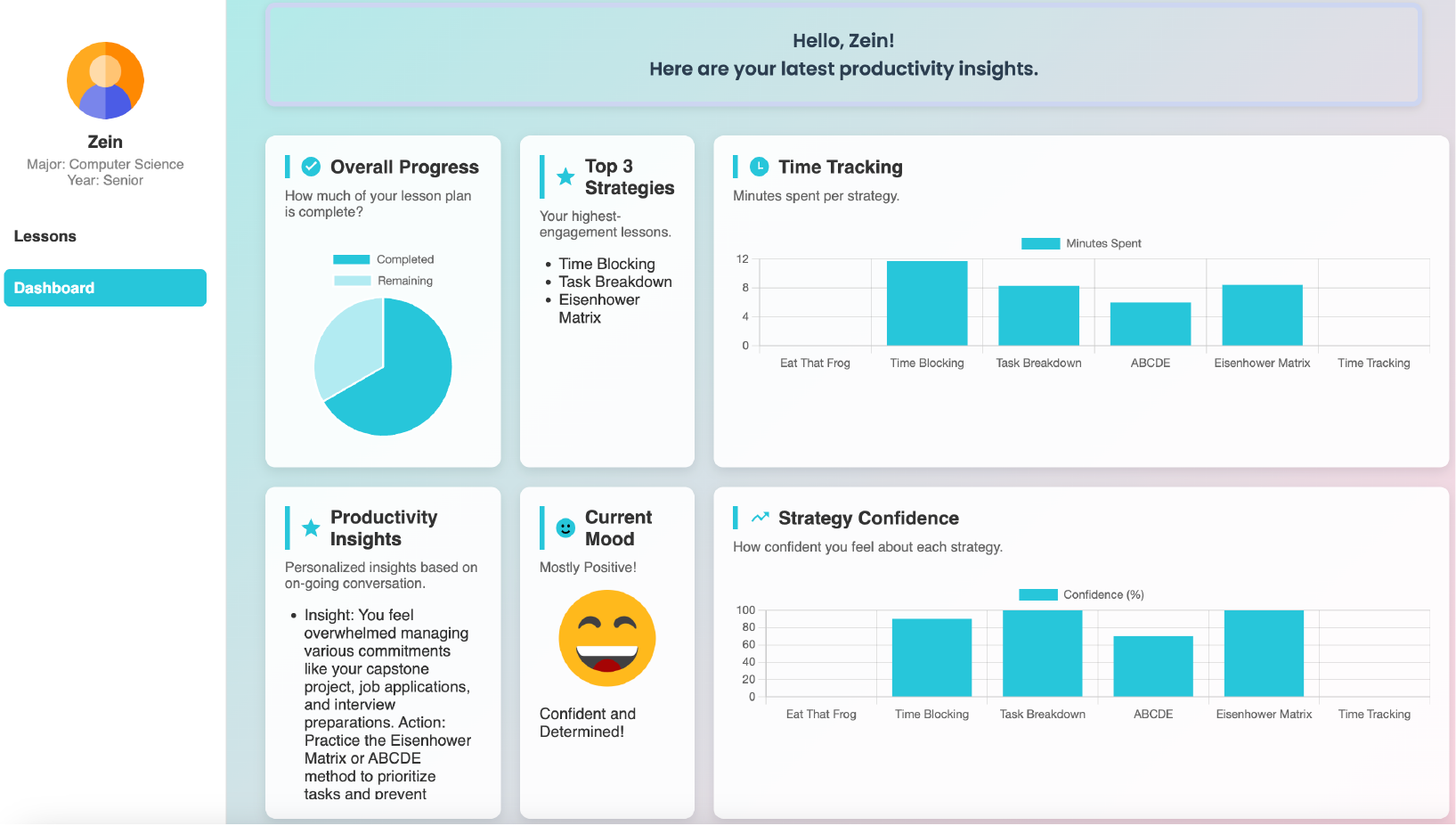}
  \caption{Dashboard Page Layout.}
  \label{fig:dashboard_page}
\end{figure}

The interactive dashboard serves as a central hub for monitoring and analyzing student progress throughout the coaching experience. Figure \ref{fig:dashboard_page} shows the layout of the dashboard. It displays the following metrics:

\begin{itemize}
    \item \textbf{Overall Progress:} Visualized as a pie chart, this metric displays the student's progress by comparing the number of lessons mastered with the remaining lessons.
    \item \textbf{Overall Mood:} The system uses prompt engineering on GPT-4 to analyze conversation data and classify the student’s mood.
    As described in the Mood Analysis Prompt (see Box~\ref{box:mood_prompt}), the LLM is instructed to categorize the overall sentiment as Positive, Negative, or Neutral, and then select a more specific description that best matches the tone of the interaction (for example, ``Confident and Determined'' for positive, ``Reflective and Observant'' for neutral, or ``Overwhelmed and Stressed'' for negative). The output is a JSON object containing both labels for display on the dashboard.
    
\begin{tcolorbox}[breakable, colframe=customcolor, colback=customcolor!30!white, coltitle=black, title=Mood Analysis Prompt, label={box:mood_prompt}, fontupper=\small]
You are an expert in sentiment analysis for productivity coaching. Based on the following conversation between a coach and a student:
[conversation\_history]
Please classify the overall sentiment of the student as one of: Positive, Negative, or Neutral.
\\ Then, select one detailed sentiment category for the student from the following lists:
\begin{itemize}
    \item Positive: [``Hopeful and Inspired'', ``Confident and Determined'', ``Energized and Focused'']
    \item Neutral: [``Calm and Grounded'', ``Reflective and Observant'', ``Balanced and Centered'']
    \item Negative: [``Overwhelmed and Stressed'', ``Frustrated and Discouraged'', ``Anxious and Uncertain'']
\end{itemize}   
Return your answer as a JSON object with two keys: ``overall'' and ``detailed''. Do not include additional text.
\end{tcolorbox}
    \item \textbf{Time Spent Per Lesson:} Presented as a bar chart, this metric shows how many minutes the student spends on each lesson, offering insights into their focus distribution.
    \item \textbf{Confidence Per Lesson:} Another bar chart displaying the student’s self-assessed confidence for each lesson. The score, recorded at the end of every lesson, helps monitor growth and identify areas that may need additional support.
    \item \textbf{Top Engagement Lessons:} For each lesson, the system evaluates text-based engagement using two metrics: word count and lexical diversity. A higher word count indicates greater engagement, while lexical diversity is used to differentiate between lessons with similar word counts. Based on these measures, the system ranks and displays the top three strategies that appear to resonate most with the student. This feature not only offers insights into which productivity methods the student finds most compelling, but also encourages reflection on which lessons may have felt less engaging or applicable.
    \item \textbf{Personal Productivity Insights:} At the end of the introductory interaction and after each lesson, the conversation history is analyzed by the LLM to generate personalized productivity insights and actionable suggestions, guiding the student toward continuous improvement.
\end{itemize}

\section{EXPERIMENTS}
\subsubsection{Setup} 
\begin{figure}
  \centering
  \includegraphics[scale=0.2]{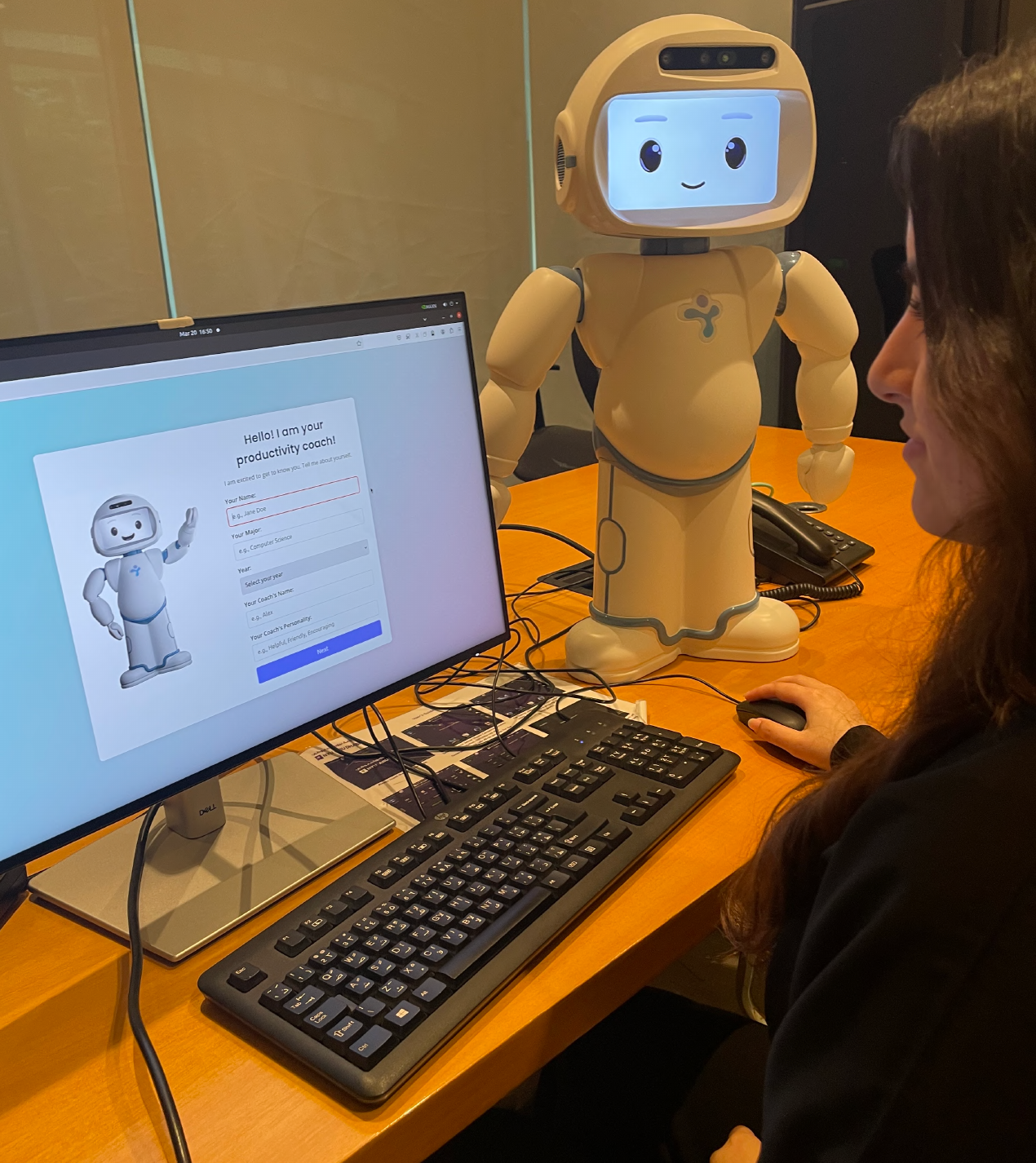}
  \caption{Experimental Setup.}
  \label{fig:setup}
\end{figure}

The study was approved by the University Institutional Review Board and conducted in a meeting room at the local institute. As shown in Figure \ref{fig:setup}, the setup included a table, chair, desktop, and QTrobot positioned diagonally across the participant and to the right of the screen.

\subsubsection{Participants} We promoted our study using flyers that had a QR code to an online sign-up form. This form collected demographic information (name, age, gender, major, and class year) and included the Executive Skills Questionnaire-Revised (ESQ-R) \cite{strait2020refinement}. We used the ESQ-R to identify students with challenges in Time Management or Plan Management, which aligned with our study's focus on productivity coaching. Only undergraduates aged 18–24 who scored above 1.5 in either area were invited to participate. In total, 15 participants (12 females and 3 males) were recruited. On average, participants scored 1.45 on the Plan Management section and 1.83 on the Time Management section.

\subsubsection{Protocol} Each experiment was conducted with the assistance of an experimenter. The protocol began with the experimenter greeting the participant and directing them to a designated chair. The study was explained, questions were answered, and the participant was prompted to complete a consent form. The experimenter also clarified that the participant could choose to complete as many lessons as they wished based on their interest and engagement. Once consent was obtained, the experimenter launched the web application and then left the room to allow the participant to work independently. The participant interacted with the robotic system. When they finished, they informed the experimenter, who returned to administer a post-experiment feedback form. After the form was completed, the experimenter thanked the participant and provided a gift voucher as compensation. Overall, each session lasted approximately 60 minutes.

\subsubsection{Surveys} At the end of the experiment, participants completed a post-experiment questionnaire. This survey incorporated the System Usability Scale (SUS) \cite{brooke1996sus} to assess the overall usability of the system. Additionally, it featured both quantitative and qualitative questions to gather feedback on every aspect of the experience, such as the coaching process, the user interface, the dashboard, the robot’s presence, its features, and its performance as a coach. This comprehensive evaluation provided valuable insights into the effectiveness and appeal of the robotic coaching system.

\section{FINDINGS}
\subsection{Participant Profiles \& Effectiveness of Lessons}
Participants reported varying levels of difficulty with time management and task prioritization. Time management was a moderately common struggle, whereas task prioritization emerged as a more significant issue. Specifically, 46\% of participants indicated they sometimes struggled with time management, while 27\% reported struggling often and another 27\% very often. In contrast, 26\% reported rarely or sometimes struggling with task prioritization, whereas the majority (67\%) struggled often and 7\% very often. 

These challenges were reflected in their evaluation of the system’s relevance and impact. Table \ref{effectiveness} displays some key results. Participants rated the coaching’s relevance and helpfulness using a 5-point Likert scale. In terms of lesson relevance, 9 out of 15 participants strongly agreed that the lessons were relevant to their needs, 5 somewhat agreed, and only 1 somewhat disagreed. The coaching was generally perceived as helpful, particularly for task prioritization: 73\% of participants rated the coaching as very helpful in this area, with an additional 13\% finding it somewhat helpful. In comparison, 67\% found the coaching somewhat helpful for time management and 20\% very helpful. 

Most of the participants did not follow the suggested lesson order, instead selecting lessons they found more interesting or unfamiliar. However, engagement with the lessons was high, as 12 out of 15 participants completed all six lessons voluntarily. Several participants attributed this to a mix of personal drive and design features of the system. P11 described the experience as driven by a ``sense of achievement of having finished everything,'' while P15 said, ``The robot’s reassurance and the interface was appealing, so I was intrigued.'' Others were motivated by the desire to improve themselves (``The desire to be more productive'') or the visual cues provided by the interface (``I saw the progress dashboard and knew how much I needed to complete so that kept me going''). P10  shared their curiosity about the robot’s adaptive responses: ``I wanted to keep interacting with the robot and see how it responds to my different struggles relating to each separate topic. And I also genuinely wanted help on those topics.'' Of the 3 participants who did not complete all lessons, 2 completed four lessons and one completed only two. All 3 attributed this to time constraints. As P13 explained, ``I couldn’t finish all the lessons in time, only 4/6 since it took me a lot of time and reflecting doing each lesson. Given more time, I would have finished everything.'' These responses suggest that both system features and personal motivation contributed meaningfully to sustained engagement with the robotic coaching system.

\begin{table}[ht]
\centering
\scriptsize
\caption{Descriptive Statistics for Lesson Relevance and Coaching Helpfulness.}
\adjustbox{max width=\columnwidth}{%
\begin{tabular}{>{\raggedright\arraybackslash}lcccc }
\toprule
\textbf{Evaluation Aspect} & \textbf{Mean $\pm$ SD} & \textbf{Median} & \textbf{Mode} \\
\midrule
\textbf{Lesson Relevance}   &  \textbf{4.40 $\pm$ 1.02}  &  \textbf{5} &  \textbf{5}      \\
\noalign{\vskip 1mm} 
\textbf{Coaching Helpfulness (Time Management) }  & \textbf{4.00 $\pm$ 0.73}  &  \textbf{4} &  \textbf{4}     \\
\noalign{\vskip 1mm} 
\textbf{Coaching Helpfulness (Task Prioritization) }  & \textbf{4.53 $\pm$ 0.88}   &  \textbf{5} &  \textbf{5}    \\

\bottomrule
\end{tabular}
}
\label{effectiveness}
\end{table}
\subsection{User Experience with the Robotic System}
Table \ref{interface_robot} summarizes participant evaluations for both interface design and the robotic coach, with more details provided in the following subsections.
\subsubsection{Interface Design and Usability}
Participants evaluated the interface using a 5-point Likert scale, ranging from ``very difficult'' to ``very easy'' for usability, and ``very unappealing'' to ``very appealing'' for visual design. Majority of the participants found the interface easy to use, with 87\% rating it as ``very easy'' to navigate and the remaining 13\% as ``somewhat easy.'' All participants reported that the navigation felt intuitive, and none encountered challenges while exploring the different components of the system. Visual appeal received more mixed feedback: nearly half (47\%) found the design ``very appealing,'' while others described it as ``somewhat appealing'' (20\%) or ``neutral'' (20\%). A few participants noted that visual consistency could be improved, suggesting ``standardized fonts and styles'' and ``a cohesive theme'' to enhance professionalism. P9 proposed the addition of motion elements, saying, ``Things that kind of pop out at you and are moving, so not everything seems frozen.'' 
Interface responsiveness was also rated on the same 5-point scale, with nearly three-quarters of participants selecting ``somewhat responsive'' or ``very responsive,'' indicating a smooth and satisfying experience overall.
\begin{table}[ht]
\centering
\scriptsize
\caption{Participant Evaluations for the Interface and Robot's Impact on Coaching.}
\adjustbox{max width=\columnwidth}{%
\begin{tabular}{>{\raggedright\arraybackslash}lcccc }
\toprule
\textbf{Evaluation Aspect} & \textbf{Mean $\pm$ SD} & \textbf{Median} & \textbf{Mode} \\
\midrule
\textbf{Interface Usability}  & \textbf{4.87 $\pm$ 0.34} &  \textbf{5} &  \textbf{5}  \\
\noalign{\vskip 1mm} 
\textbf{Interface Aesthetics}  & \textbf{4.00 $\pm$ 1.09}  &  \textbf{4} &  \textbf{5}     \\
\noalign{\vskip 1mm} 
\textbf{Interface Responsiveness}  & \textbf{4.13 $\pm$ 0.96}   &  \textbf{4} &  \textbf{5}     \\
\noalign{\vskip 1mm} 
\textbf{Robot’s Contribution to Coaching Effectiveness}  & \textbf{3.40 $\pm$ 1.50}  &  \textbf{4} &  \textbf{5}      \\
\noalign{\vskip 1mm} 
\textbf{Robot’s Contribution to Coaching Engagement}  & \textbf{3.67 $\pm$ 1.53} &  \textbf{4} &  \textbf{5}        \\
\bottomrule
\end{tabular}
}
\label{interface_robot}
\end{table}
\subsubsection{Perceptions of the Robotic Coach}
\begin{figure}
  \centering
  \includegraphics[scale=0.48]{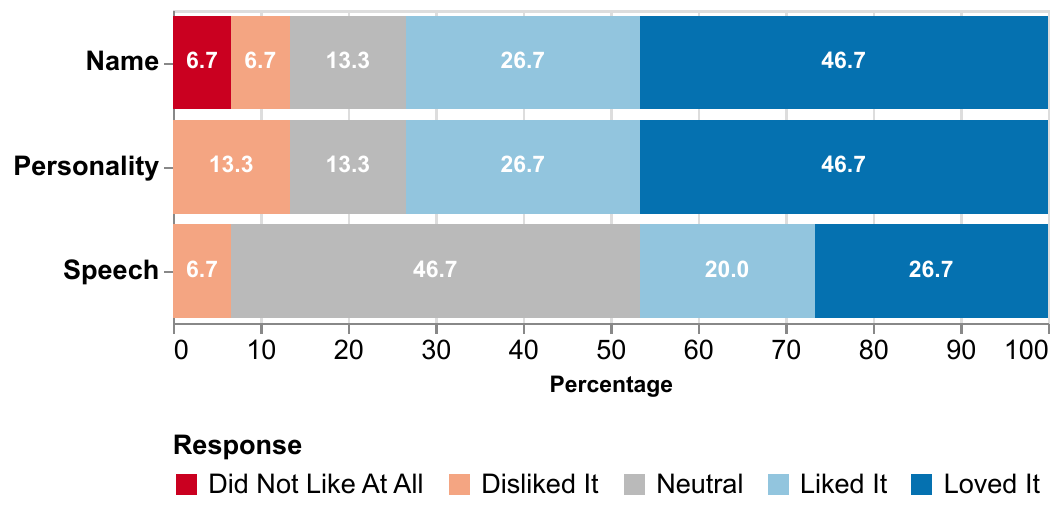}
  \caption{Participants' Evaluation of the Robot’s Personalization Features using a 5-Point Likert Scale.}
  \label{fig:personalization}
\end{figure}

\begin{itemize}
    \item \textbf{Perceived Presence and Support.} The robot played a central role in shaping participants’ engagement with the coaching experience. On a 5-point Likert scale, 53\% of participants somewhat or strongly agreed that its presence made the coaching more effective, and 60\% felt it helped them stay engaged throughout the session. Beyond these ratings, many described how the robot’s physical presence fostered a sense of social connection and accountability, elevating the experience beyond what a purely text-based system could offer. P12 described the robot's presence to be ``captivating and allowed me for a better understanding of the task. It simulated a person standing next to me while I was reading a response in chat.'' P14 shared similar views by saying, ``I would have felt less motivated if I was just reading text off the screen. Having something beside me and interacting with me made me feel like I needed to engage more.'' The robot was also appreciated for providing a lower-pressure environment compared to a human coach: ``It was better for me to interact with robot without having to necessarily worry about how a human coach would react.'' For many, its expressive gestures and friendly tone added to the sense of presence and relatability, with one participant noting, ``It was very cute and its answers were direct but said in a nice way, like a friend giving you advice.'' Together, these elements made the robot a supportive and motivating coach that helped participants stay attentive and immersed in the session.
    \item \textbf{Speech and Gestures.} Experiences with the robot’s speech and gestures were more varied. Many participants appreciated the gestures, describing them as expressive and helpful for maintaining attention. However, some found them distracting or unnecessary. The mechanical sounds during movement were noted by a few as disruptive. Speech, on the other hand, was seen as a key engagement tool. It was described as cheerful and clear, though some participants felt it was too slow or lacked natural flow. Despite these critiques, most participants felt that the combination of verbal delivery and nonverbal cues helped make the interaction feel more dynamic and human-like.
    \item \textbf{Personalization Features.} As shown in Figure \ref{fig:personalization}, customization features, such as naming the robot, defining its personality, and toggling its speech, were generally well-received. Most participants appreciated the \textit{ability to name the robot}, giving the feature an average rating of 4 out of 5. Many chose names with personal significance. P11 named the robot after their sister, explaining, ``She constantly reminds me to do my work.'' P13 chose ``Zangar,'' their brother’s name, saying, ``I felt a connection to it and wanted to listen to it.'' Others opted for names that made the robot feel more approachable or emotionally resonant, such as ``Roberta... I also think having a female name made it more likeable for me,'' and ``Kabour, it made me feel safe to talk to it.'' These examples suggest that name customization, though simple, added a meaningful layer of familiarity and comfort to the interaction.

    Participants were also able to define \textit{personality traits for the robot}, such as friendly, helpful, or strict. The feature received an average rating of 4.07 out of 5. Many felt their choices shaped the tone of the session in a positive way. P12 shared, ``I wanted it to be strict and helpful so the robot was explaining strategies to me without offering compromises. It made the experience more enjoyable.'' Others found that selecting a friendly personality ``made the mood of the session better for sure'' and ``helped my experience because it was encouraging and validating.'' However, not all participants noticed a strong impact from this feature; 2 participants felt the personality selection had little influence on the robot’s behavior during the session.

    \hspace*{1em} Participants also valued the \textit{speech toggle} for its flexibility, giving it an average rating of 3.67. Some preferred keeping the speech on as it helped them maintain focus and feel more immersed (``The speech helped me to focus more'' and ``I prefer listening to the robot’s speech while reading as it gives me a sense of presence next to me''). Others opted to turn it off when it became distracting or felt too slow. Overall, these personalization features allowed users to adapt the system to their preferences, enhancing both comfort and engagement.
\end{itemize}
\subsubsection{Dashboard Metrics}
\begin{figure}
  \centering
  \includegraphics[scale=0.41]{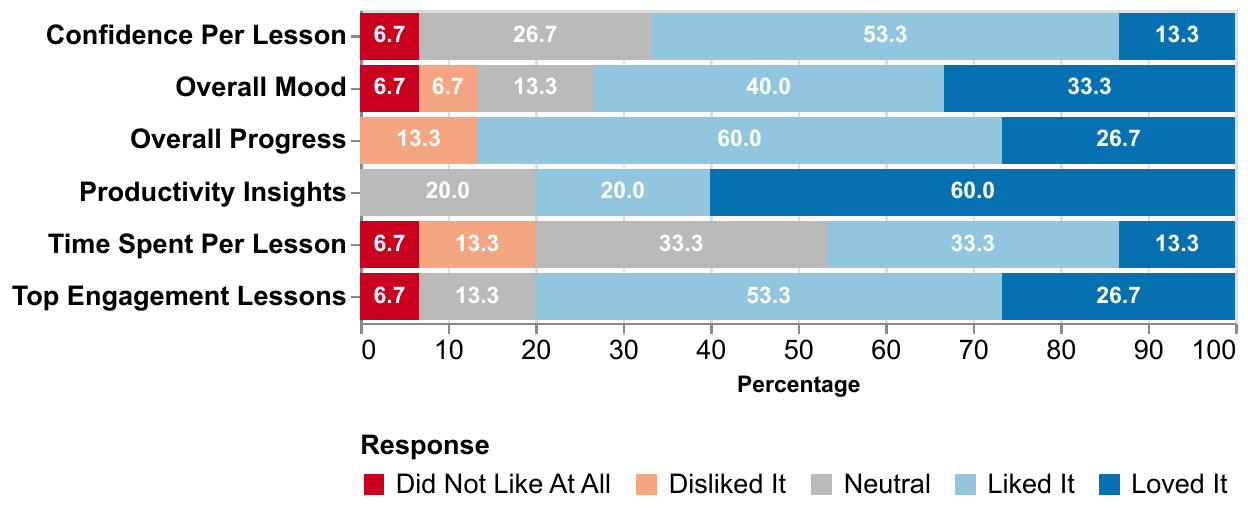}
  \caption{Participants' Evaluation of Dashboard Elements on a 5-Point Likert Scale.}
  \label{fig:dashboard}
\end{figure}

The dashboard was another important component of the system that was highly appreciated. 13 out of 15 participants found it somewhat or very helpful. As shown in Figure \ref{fig:dashboard}, all elements were rated positively. Participants described it as both motivating and reflective, helping them track progress and better understand their productivity patterns. P8 shared, ``The dashboard was super helpful in defining my goals and progress which helped motivate me to keep going,'' while P1 noted, ``I discovered some things I didn’t know, the dashboard showed how eager I am to start on solving my procrastination issues.''

Beyond progress tracking, participants also found the dashboard useful for reflecting on which strategies worked for them and how they engaged with different lessons. P15 remarked, ``It was nice to see what worked for me and how long it took me to understand the methods,'' while P2 said, ``It showed me what strategies would be effective for me and what I found to be useful and happy to engage with.''

Among all elements, the personal productivity insights were described as the most meaningful. P13 shared, ``Mostly I was looking at the profile insights at the left bottom of the dashboard page, since this is what I wanted to learn about myself through coaching.'' Others appreciated how their inputs were interpreted and synthesized: ``It talked about my preferences and the different types of information that I was giving it, and it interpreted it and put it all in one place. This made it easier to be objective about myself and realize some things about me that I was not fully aware of beforehand.'' Another participant simply described it as ``a great resource to refer to.''

Although the general feedback was positive, participants offered constructive suggestions. Some questioned how engagement was calculated, with P8 noting, ``There could be more explanation on the top 3 engagement lessons category because from what I understood it was just whichever sections I spent the longest on but I don’t think that’s really an indicator of my engagement level.'' Others suggested design improvements, recommending the dashboard be ``more colourful and engaging,'' with sections ``spread out then we can see everything clearly,'' and the personal insights area ``put in a separate place with bigger text.'' Participants also proposed additional metrics, such as displaying more solutions for the user, including line graphs showing how much they liked each lesson, and summarizing what worked or did not through success/failure rates.

\subsubsection{System Usability and Overall Coaching Experience}
\begin{figure}[ht]
    \centering
    \begin{subfigure}[b]{0.47\columnwidth}
        \centering
         \captionsetup{justification=raggedright, singlelinecheck=false, labelfont=bf, labelsep=period, font=small} 
        \caption{}
        \includegraphics[width=\columnwidth]{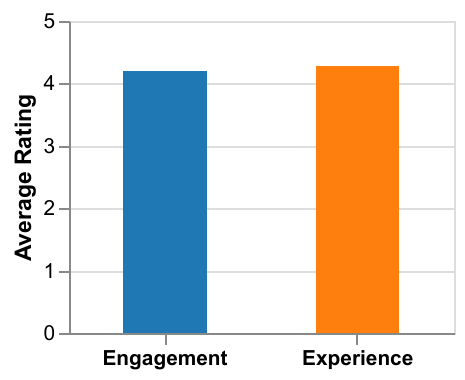} 
        \label{fig:overall}
    \end{subfigure}
    \hfill
    \begin{subfigure}[b]{0.47\columnwidth}
        \centering
         \captionsetup{justification=raggedright, singlelinecheck=false, labelfont=bf, labelsep=period, font=small} 
        \caption{}
        \includegraphics[width=\columnwidth]{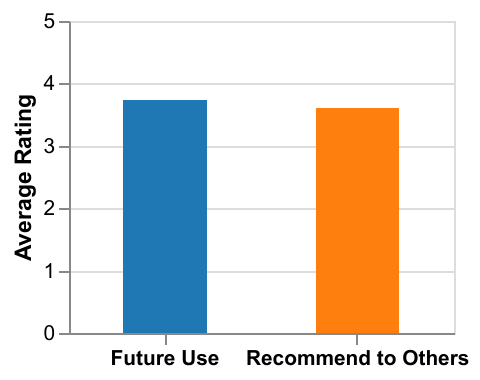} 
        \label{fig:future_use}
    \end{subfigure}

    \caption{Participants’ Average Ratings: (A) Overall Experience \& Engagement with the Robotic System and (B) Likelihood of Future Use \& Recommendation.}
    \label{fig:overall_coaching}
\end{figure}

Participants generally had a positive experience with the coaching system. The SUS score was 79.2, well above the average of 68. This corresponds to an ``A-'' grade, indicating high usability based on established benchmarks \cite{lewis2018item}. As shown in Figure \ref{fig:overall}, on average, participants rated their overall experience as 4.3 out of 5, and found the session engaging with an average rating of 4.2.

Participants appreciated how the coaching process encouraged meaningful self-reflection. As P13 shared, ``I liked that the coaching allowed me to take a step back and really reflect on the tasks that I am doing right now. To understand where I am right now, and what steps should I take next.'' The coach was also perceived as attentive and goal-oriented, which helped build trust and sustain engagement. Additionally, several participants highlighted how the system made the experience feel more personalized by adapting responses to their context. For instance, P10 recalled, ``I was pleasantly surprised when at one point it used a physics analogy to help me understand better,'' while P9 noted, ``It was very well informed about things that I mentioned about myself, without me giving an prior knowledge, like how me mentioning doing theatre allowed it to deduce that learning lines and preparing for auditions and rehearsals are important things.''

However, a few participants noted limitations in both content and interaction. P2 described the advice as ``a bit theoretical,'' and wished for more practical, step-by-step guidance. Others found typing responses tiring, which impacted their engagement. As P8 explained, ``I found it tiresome to type out my responses which ultimately made me less engaged because I felt lazy to fully express my thoughts.''

Participants were also asked to compare the system to human coaching using a 5-point Likert scale. With human coaching being considered the gold standard, 9 out of 15 participants rated the robotic coaching as somewhat below that level, 4 found it somewhat better, and 2 rated it as the same. This suggests a moderate but not substantial gap in perceived effectiveness of the robotic system. Participants also expressed moderate interest in continued use and recommendation of the system, with average ratings of 3.7 out of 5 for future use and 3.6 out of 5 for recommending it to others (see Figure \ref{fig:future_use}).

\section{CONCLUSIONS}
This work serves as a proof-of-concept for using socially assistive robots as educational productivity coaches in a higher education setting. Our study demonstrates that a SAR-based coaching system can offer meaningful support in areas such as skill building, self-awareness, and motivation to college students navigating academic challenges. By integrating lessons on time management and task prioritization with an interactive dashboard that delivers personalized productivity insights, the system achieved strong usability and high engagement. However, there are several limitations. The current study had a higher proportion of female participants, which may influence the generalizability of the findings. Additionally, speech-to-text functionality for user input could not be integrated due to current limitations in existing automatic speech recognition systems. Future work will focus on incorporating participant feedback, doing longitudinal studies, and testing the system with a more diverse student population. We also aim to expand this proof-of-concept into a more flexible and adaptive coaching system, capable of addressing a wider range of productivity-related challenges.

\addtolength{\textheight}{-12cm}   




\section*{ACKNOWLEDGMENT}

This work is supported in part by the NYUAD Center for Artificial Intelligence and Robotics, funded by Tamkeen under the NYUAD Research
Institute Award CG010.


\bibliographystyle{IEEEtran}
\bibliography{sample-base}

\end{document}